\documentclass{article}

\usepackage[final]{neurips_2020}

\usepackage[utf8]{inputenc} 
\usepackage[T1]{fontenc}    
\usepackage[hidelinks]{hyperref}       
\usepackage{url}            
\usepackage{booktabs}       
\usepackage{amsfonts}       
\usepackage{nicefrac}       
\usepackage{microtype}      
\usepackage{graphicx}
\usepackage{todonotes}
\usepackage{subcaption}
\usepackage{amsmath}
\usepackage{bbm}
\usepackage[ruled,vlined]{algorithm2e}
\usepackage{enumitem}
\usepackage{setspace}
\usepackage{float}
\usepackage{wrapfig}
\usepackage[hang,flushmargin]{footmisc} 
\usepackage{caption} 

\title{Influence-Augmented Online Planning for Complex Environments}

\author{Jinke He \\
        Department of Intelligent Systems \\
        Delft University of Technology \\
        \texttt{J.He-4@tudelft.nl}
        \And
        Miguel Suau \\
        Department of Intelligent Systems \\
        Delft University of Technology \\
        \texttt{M.SuaudeCastro@tudelft.nl}
        \And
        Frans A. Oliehoek \\
        Department of Intelligent Systems \\
        Delft University of Technology \\
        \texttt{F.A.Oliehoek@tudelft.nl}
}

\begin{document}

\maketitle


\newcommand{\E}{\mathop{\mathbb{E}}}

\renewcommand{\S}{\mathcal{S}} 
\newcommand{\A}{\mathcal{A}} 
\newcommand{\T}{\mathcal{T}} 
\renewcommand{\O}{\mathcal{O}} 
\newcommand{\R}{\mathcal{R}} 
\renewcommand{\H}{\mathcal{H}}

\newcommand{\local}{\mathit{local}}
\newcommand{\M}{\mathcal{M}}
\newcommand{\G}{\mathcal{G}}
\newcommand{\GM}{\mathcal{M}_\mathtt{global}}
\newcommand{\IALM}{\mathcal{M}_\mathtt{IALM}}
\newcommand{\SL}{\mathcal{S}^\mathtt{IALM}}
\newcommand{\TL}{\mathcal{T}^\mathtt{IALM}}
\newcommand{\bL}{b_0^\mathtt{IALM}}

\newcommand{\simGM}{\mathcal{G}_\mathtt{global}}
\newcommand{\simIALM}{\mathcal{G}_\mathtt{IALM}}
\newcommand{\simIALMp}{\mathcal{G}_\mathtt{IALM}^\theta}

\newcommand{\Slocal}{S^{\mathtt{local}}}
\newcommand{\slocal}{s^{\mathtt{local}}}
\newcommand{\Snonlocal}{S^{\neg \mathtt{local}}}
\newcommand{\snonlocal}{s^{\neg \mathtt{local}}}
\newcommand{\SIALM}{S^{\mathtt{IALM}}}
\newcommand{\sIALM}{s^{\mathtt{IALM}}}

\newcommand{\D}{\mathcal{D}} 
\newcommand{\piexplore}{\pi_{\mathtt{explore}}}

\newcommand{\KLD}[2]{D_{\mathrm{KL}} \left( \left. \left. #1 \right|\right| #2 \right) }

\newcommand{\aip}{\hat{I}_\theta}

\newcommand{\Ysrc}{Y^{\mathtt{src}}}
\newcommand{\Xdest}{X^{\mathtt{dest}}}
\newcommand{\ysrc}{y^{\mathtt{src}}}
\newcommand{\xdest}{x^{\mathtt{dest}}}
\newcommand{\Xnondest}{X^{\neg \mathtt{dest}}}
\newcommand{\xnondest}{x^{\neg \mathtt{dest}}}

\newcommand{\simLocal}{\mathcal{G}_{\mathtt{local}}}

\newcommand{\SIALS}{S^{\mathtt{IALS}}}
\newcommand{\sIALS}{s^{\mathtt{IALS}}}


\newcommand{\todoinline}[1]{\todo[inline,color=white]{Jinke: #1}}

\newcommand{\Miguel}[1]{\todo[inline,color=orange]{Miguel: #1}}
\newcommand{\Frans}[1]{\todo[inline,color=blue]{Frans: #1}}

\begin{abstract}
    How can we plan efficiently in real time to control an agent in a complex environment that may involve many other agents? While existing sample-based planners have enjoyed empirical success in large POMDPs, their performance heavily relies on a fast simulator. However, real-world scenarios are complex in nature and their simulators are often computationally demanding, which severely limits the performance of online planners. In this work, we propose influence-augmented online planning, a principled method to transform a factored simulator of the entire environment into a local simulator that samples only the state variables that are most relevant to the observation and reward of the planning agent and captures the incoming influence from the rest of the environment using machine learning methods. Our main experimental results show that planning on this less accurate but much faster local simulator with POMCP leads to higher real-time planning performance than planning on the simulator that models the entire environment.
\end{abstract}

\section{Introduction}
\label{sec:Introduction}

We consider the online planning setting where we control an agent in a complex
environment that is partially observable and may involve many other agents.
When the policies of other agents are known, the entire environment can be modeled
as a Partially Observable Markov Decision Process (POMDP) \citep{Kaelbling1998}, and traditional online planning approaches can
be applied. While sample-based planners like POMCP \citep{Silver2010} have been
shown effective for large POMDPs, their performance relies heavily on a fast
simulator to perform a vast number of Monte Carlo simulations in a step.
However, many real-world scenarios are complex in nature, making simulators
that capture the dynamics of the entire environment extremely computationally
demanding and hence preventing existing planners from being useful in practice.
Towards effective planning in realistic scenarios, this work is motivated by
the question: can we significantly speed up a simulator by replacing the part of the environment that is less important with an approximate learned model?

\begin{figure}
    \centering
    \begin{subfigure}[b]{.43\textwidth}
        \centering
        \includegraphics[scale=0.645]{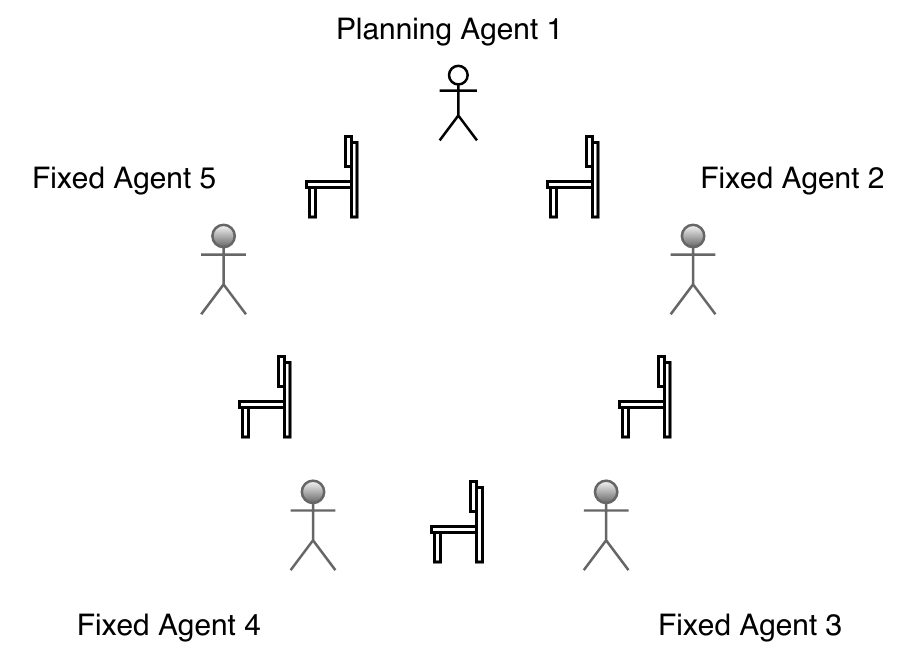}
        \caption{}
        \label{fig:5_agent_gac}
    \end{subfigure}
    \begin{subfigure}[b]{.55\textwidth}
        \centering
        \includegraphics[scale=0.018]{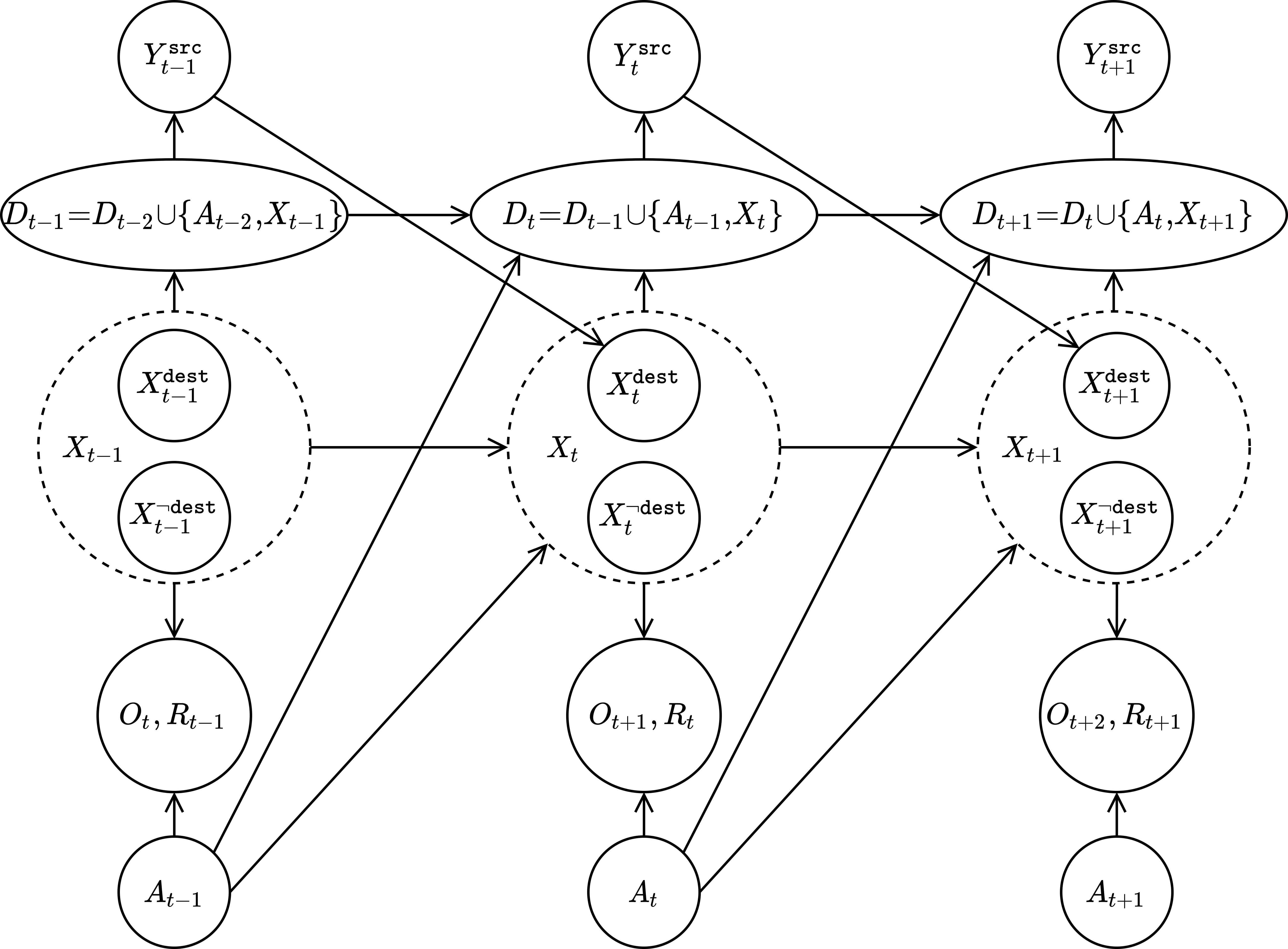}
        \caption{}
        \label{fig:DBN-IALM}
    \end{subfigure}
    \caption{Left: Controlling a single agent in the Grab A Chair game with $4$ other agents. Right: Dynamic Bayesian Network for the influence-augmented local model.}
\end{figure}

We build on the multi-agent decision making literature that tries to identify compact representations of complex environments for an agent to make optimal decisions \citep{Becker2003,Becker2004,Petrik2009,Witwicki2010}. These methods exploit the fact that in many structured domains, only a small set of (state) variables, which we call \textit{local (state) factors}, of the environment directly affects the observation and reward of the agent. The rest of the environment can only impact the agent indirectly through their influence on the local factors. For example, Figure~\ref{fig:5_agent_gac} shows a game called Grab A
Chair, in which there are $N$ agents that, at every time step, need to decide whether they will try to grab the chair on their left or right side. An agent can only secure a chair if that chair is not targeted by the other neighboring agent. At the end of every step, each agent only observes whether it obtains the chair, without knowing the decisions of others. Additionally, there is a noise on observation, i.e., a chance that the agent gets an incorrect observation. In this game, it is clear that to the planning agent, whose goal is to obtain a chair at as many steps as possible, 
the decisions of neighboring agents 2 and 5 are more important than those of
agents 3 and 4 as the former directly determine if the planning agent can
secure a chair. In other words, only agents 2 and 5 directly \emph{influence} agent 1's local decision making, while 
agents 3 and 4 may only do so indirectly.

To utilize this fact, we propose influence-augmented online planning, a principled method that transforms a factored simulator of the entire environment, called \emph{global simulator}, into a faster \emph{influence-augmented local simulator (IALS)}. The IALS simulates only the local factors, and concisely captures the influence of the external factors by predicting only the subset of them, called \emph{source factors}, that directly affect the local factors. Using off-the-shelf supervised learning methods, the influence predictor is learned offline with data collected from the global simulator. Our intuition is that when planning with sample-based planners, the advantage that substantially more simulations can be performed in the IALS may outweigh the simulation inaccuracy caused by approximating the incoming influence. In this paper, we investigate this hypothesis, and show that this approach can indeed lead to improved online planning performance. 

In detail, our planning experiments with POMCP show that, by replacing the global simulator with an IALS that learns the incoming influence with a recurrent neural network (RNN), we achieve matching performance while using much less time. More importantly, our real-time online planning experiments show that planning with the less accurate but much faster IALS yields better performance than planning with the global simulator in a complex environment, when the planning time per step is constrained. In addition, we find that learning an accurate influence predictor is more important for good performance when the local planning problem is tightly coupled with the rest of the environment.

\section{Background}
\label{sec:background}


\subsection{POMDP}

A Partially Observable Markov Decision Process (POMDP) \citep{Kaelbling1998} models the interactive process of an agent making decisions and receiving feedback in an environment with limited observation. Formally, a POMDP is a tuple $\M = (\S, \A, \T, \R, \Omega, \O, b_0, \gamma)$ where $\S$, $\A$, $\Omega$ are the set of environment states, actions and observations. The transition function $\T: \S \times \A \rightarrow \Delta(S)$ determines the distribution over the next state $S_{t+1}$ given the previous state $S_t$ and action $A_t$, where $\Delta(S)$ denotes the space of probability distributions over $S$. On transition, the agent receives a reward $R_{t} \sim \R(S_{t+1}, A_t)$ and a new observation $O_{t+1} \sim \O(S_{t+1}, A_t)$. A policy $\pi$ is a behavioral strategy that maps an action-observation history $h_t = \{s_0, o_1, \ldots, a_{t-1}, o_t\}$ to a distribution over actions. The belief state $b_t \in \Delta(S)$ is a sufficient statistic of the history $h_t$, representing the distribution over $S_t$ conditioned on $h_t$, with $b_0$ being the initial belief and known. The value function $V^\pi(h_t)$ measures the expected discounted return from $h_t$ by following $\pi$ afterwards, $V^\pi(h_t) = \E_{\pi} [\sum_{k=0}^\infty \gamma^{k} R_{t+k}| H_t = h_t]$, where $\gamma \in [0,1]$ is the discount factor, with the optimal value function $V^*(h_t) = \max_\pi V^\pi(h_t)$ measuring the maximally achievable value from $h_t$. The optimal value of a POMDP $\M$ is defined as $V^*_\M = V^*(b_0) = \max_{\pi} \E_{\pi} [\sum_{t=0}^\infty \gamma^t R_{t}]$.




In structured domains, the state space $\S$ of a POMDP can be factorized into a finite set of state variables $\S = \{\S^1, \ldots, \S^N\}$, whose conditional independence between each other and the observation and reward variables can be utilized to construct a more compact representation of the POMDP called Dynamic Bayesian Network (DBN) \citep{Boutilier1999}. For convenience, we use the notation $S_t$ to refer to both the set of state variables and the joint random variable over them.

\subsection{Sample-based Online Planning in POMDPs}
\label{background:POMCP}
Many real-world decision making problems are so complex that finding a policy that performs well in all situations is not possible. In such cases, online planning methods which aim to find a local policy $\pi(\cdot | h_t)$ that maximizes $V^\pi(h_t)$ when observing a history $h_t$ can lead to better performance. In fully observable case, sample-based planning methods that evaluate actions by performing sample-based lookahead in a simulator have been shown effective for large problems with sample complexity irrelevant to the state space size \citep{Kearns}. Monte Carlo Tree Search (MCTS) is a popular family of sample-based planning methods \citep{Coulom2006,Kocsis,Browne2012} that implement a highly selective search by building a lookahead tree and focusing the search on the most promising branches during the planning process.

POMCP proposed by \citet{Silver2010} extends MCTS to large POMDPs, addressing both the curse of dimensionality and the curse of history with Monte Carlo simulation in a generative simulator $\G$ that samples transitions. To avoid the expensive Bayesian belief update, POMCP approximates the belief state with an unweighted particle filter. Similar to MCTS, POMCP maintains a lookahead tree with  nodes representing the simulated histories $h$ that follow the real history $h_t$. To plan for an action, POMCP repeatedly samples states from the particle pool $B(h_t)$ at the root node. By simulating a state to the end, with actions selected by the UCB1 algorithm \citep{Auer2002} inside the tree and a random policy during the rollout, the visited nodes are updated with the simulated return and the tree is expanded with the first newly encountered history. When the planning terminates, POMCP executes the action $a_t$ with the highest average return and prunes the tree by making the history $h_{t+1}{=}h_t a_t o_{t+1}$ the new root node. Notably, POMCP shares the simulations between tree search and belief update by maintaining a pool of encountered particles in every node during the tree search. This way, when $h_{t+1}$ is made the new root node, $B(h_{t+1})$ becomes the new estimated belief state.

\subsection{Influence-Based Abstraction}
\label{background:IBA}
Influence-Based Abstraction (IBA) \citep{Oliehoek2012} is a state abstraction method \citep{Li} which abstracts away state variables that do not directly affect the observation and reward of the agent, without a loss in the value. In the following, we provide a brief introduction on IBA and refer interested readers to \citet{Oliehoek2019} for more details.

Given a factored POMDP, which we call the global model $\GM = (\S, \A, \T, \R, \Omega, \O, b_0, \gamma)$, IBA splits the set of state variables that constitute the state space $\S$ into two disjoint subsets, the set of \emph{local state variables} $X$ that include at least the parent variables of the observation and reward variables and the set of \emph{non-local state variables} $Y = \S \backslash X$. 

IBA then defines an influence-augmented local model (IALM)  $\IALM$,
where the non-local state variables $Y$ are marginalized out. To define the transition function $\TL$ on only the local state
variables $X$, IBA differentiates the local state variables $\Xdest \subseteq X$ that are directly affected by the non-local state variables, called \emph{influence destination state variables}, from those that are not $\Xnondest = X \backslash \Xdest$. In addition, IBA defines the non-local state variables $\Ysrc \subseteq Y$ that directly affect $X$ as \emph{influence source state variables}. In other words, the non-local state variables $Y$ influences the local state variables $X$ only through $\Ysrc$ affecting $\Xdest$ as shown in Figure \ref{fig:DBN-IALM}. Since abstracting away $Y_0,
\ldots, Y_{t-1}$ creates a dependency of $\Ysrc_t$ on the history of local
states and actions, the state $\SIALM_t$ needs to include both the local state
$X_t = (\Xnondest_t, \Xdest_t)$ and the so-called d-separation set $D_t$, which encodes the relevant parts of the local history. Given this, $\TL$ is defined as follows:
\begin{align*}
    \centering
    &\TL(\SIALM_{t+1} | \SIALM_t, A_t) = \Pr(X_{t+1}, D_{t+1}| X_t, D_t, A_t) \\
    =& \Pr(\Xnondest_{t+1} | X_t, A_t) \mathbbm{1}(D_{t+1} = d(X_t,A_t,X_{t+1},D_t))\Pr(\Xdest_{t+1} | X_t, D_t, A_t) \\
    =& \Pr(\Xnondest_{t+1} | X_t, A_t) \mathbbm{1}(D_{t+1} = d(X_t,A_t,X_{t+1},D_t)) \sum_{\ysrc_t} I(\ysrc_t | D_{t}) \Pr(\Xdest_{t+1} | X_t, \ysrc_t, A_t) 
\end{align*}
where $\mathbbm{1}(\cdot)$ is the indicator function and the notation $I$ is introduced as the influence predictor, $I(\ysrc_t | D_{t}) = \Pr(\ysrc_t | D_{t})$. The function $d$ selects those variables that are relevant to predict the influence sources $Y_t^{src}$. In this paper, we set $d(X_t,A_t,X_{t+1},D_t)) = D_t \cup \{A_{t-1}, X_t\}$. That is, even though in general it is possible to condition on the history of a subset of local states and actions, we just use the entire history of local states and actions for simplicity (see \cite{suau2019influence} for an exploitation of this aspect of IBA in the context of Deep RL). The IALM is then
formally defined as $\IALM = (\SL, \A, \TL, \R, \Omega, \O, b_0, \gamma)$ where
the observation function $\O$, reward function $\R$ and the initial belief $b_0$
remain unchanged because of the definition of the local state variables $X$.
Theorem 1 in \citet{Oliehoek2019} proves that this is a lossless abstraction by
showing the optimal value of the IALM matches that of the global model, $V^*_{\IALM} = V^*_{\GM}$.

\section{Influence Augmented Online Planning}
\label{sec:method}

While IBA results in an IALM $\IALM$ that abstracts away non-local state variables $Y$ in a lossless way, it is not useful in practice because computing the distribution $I(\Ysrc_t|D_t)$ exactly is in general intractable. Our approach trades off between the time spent before and during the online planning, by approximating $I(\Ysrc_t|D_t)$ with a function approximator $\aip$ learned offline. The learned influence predictor $\aip$ will then be integrated with an accurate local simulator $\simLocal$ to construct an influence-augmented local simulator (IALS) that only simulates the local state variables $X$ but concisely captures the influence of the non-local state variables $Y$ by predicting the influence source state variables $\Ysrc$ with $\aip$. During the online planning, the integrated IALS will be used to replace the accurate but slow global simulator to speed up the simulations for the sample-based online planners.

Our motivation is that by simulating the local transitions that directly decide the observation and reward of the agent with an accurate local simulator, the simulation inaccuracy caused by approximating the distribution $I(\Ysrc_t|D_t)$ with $\aip$ can be overcome by the advantage that simulations can be performed significantly faster in the IALS, which is essential to sample-based planners like POMCP \citep{Silver2010}, leading to improved online planning performance in realistic scenarios with limited planning time. Our overall approach, influence-augmented online planning, is presented in Algorithm \ref{algo:IAOP}, followed by our method to learn an approximate influence predictor with recurrent neural networks (RNNs) \citep{Hochreiter1997,Cho2014} and integrate it with a local simulator to form a plannable IALS for sample-based planners.
\begin{algorithm}
    \setstretch{1.02}
    \SetAlgoLined
    \SetKwProg{offline}{Offline}{}{}
    \SetKwProg{online}{Online}{}{}
    \SetKwInOut{Input}{input}
    \Input{a real environment $\mathtt{env}$}
    \Input{a global simulator $\simGM$ and a local simulator $\mathcal{G}_{\mathtt{local}}$}
    \Input{an exploratory policy $\pi_{\mathtt{explore}}$}
    \Input{a sample-based planner $\mathtt{planner}$ with a termination condition $T$, e.g., a fixed time limit}
    \Input{a planning horizon $\H$}
    \offline{ Influence Learning}{
        Collect a dataset $\mathcal{D}$ of input sequences $D_{\H-1}{=}(A_{i-1}, X_i)_{i=1}^{\H-1}$ and target sequences $(\Ysrc_i)_{i=1}^{\H-1}$ by interacting with the global simulator $\simGM$ using the policy $\pi_{\mathtt{explore}}$\;
        Train an approximate influence predictor $\aip$ on the dataset $\mathcal{D}$ by minimizing the average empirical KL Divergence between $I(\cdot | D_t)$ and $\aip(\cdot | D_t)$ \;
    }
    \online{ Planning with a sample-based planner} {
        Integrate the local simulator $\simLocal$ and the learned influence predictor $\aip$ into an IALS $\simIALMp$\;
        \For{$t=0, \ldots, \H{-}1$}{
            plan for an action until $T$ is met: $a_t = \mathtt{planner.plan}(\simIALMp, T)$\;
            execute the action in the real environment: $o_{t+1} = \mathtt{env.act}(a_t)$ \;
            process the new observation: $\mathtt{planner.observe}(o_{t+1})$
        }
    }
    \caption{Influence-Augmented Online Planning}
    \label{algo:IAOP}
\end{algorithm}
\subsection{Learning Approximate Influence Predictor Offline with RNNs}
The dependency of $I(\Ysrc_t|D_t)$ on the d-separation set $D_t$ renders it infeasible to be computed exactly online or offline. In this work we learn an approximate influence predictor offline with RNNs by formalizing it as a supervised sequential classification problem. 

For planning with horizon $\H$, we need to predict the conditional distribution over the influence source state $I(\Ysrc_t|D_t)$ for $t=1$ to $\H{-}1$. We do not need to predict $I(\Ysrc_0|D_0)$ as it is the initial belief over the influence source state. As RNNs require the input size to be constant for every time step, we drop the initial local state $X_0$ from $D_t$ so that the input to RNNs at time step $t$ is $\{A_{t-1}, X_t\}$ and the target is $\Ysrc_t$. If there exists a distribution from which we can sample a dataset $\mathcal{D}$ of input sequence $D_{\H-1}$ and target sequence $(\Ysrc_1, \ldots, \Ysrc_{\H_1})$, then this is a classic sequential classification setup that can be learned by training a RNN $\aip$ to minimize the average empirical KL divergence between $I(\cdot | D_t)$ and $\aip(\cdot | D_t)$ with stochastic gradient descent (SGD) \citep{Ruder2016}, which yields a cross-entropy loss in practice. While we leave the question on how can we collect the dataset $\mathcal{D}$ in a way that maximizes the online planning performance for future investigation, in this paper we use a uniform random policy to sample $\mathcal{D}$ from the global simulator $\simGM$.

\subsection{Integrating the Local Simulator and RNN Influence Predictor for Online Planning}
To plan online in a POMDP, sample-based planners like POMCP \citep{Silver2010} require a generative simulator that supports sampling the initial states and transitions. As shown in Figure \ref{fig:DBN-IALM}, to sample a transition in the IALS $\simIALMp$, we need to first sample an influence source state $\Ysrc_t$ and then sample the local transitions in the local simulator $\simLocal$. While in the original formulation of IBA, $\aip(\Ysrc_t|D_t)$ conditions on the d-separation set $D_t$ which grows with actions $A_t$ and new local states $X_{t+1}$ at every time step, we avoid feeding the entire $D_t$ into RNNs for every prediction of $\Ysrc_t$ by taking the advantage of RNNs whose hidden state $Z_t$ is a sufficient statistic of the previous inputs. As a result, we use $\SIALM_t = (X_t, \Ysrc_t, Z_t)$ as the state of the IALS in practice. The transition $s_{t+1}^{\mathtt{IALM}}, o_{t+1}, r_{t+1} \sim \simIALM^\theta({s_{t}^{\mathtt{IALM}},a_{t}})$ can then be sampled in two steps:
\begin{itemize}[leftmargin=12pt,noitemsep,topsep=0pt]
    \setlength\itemsep{0em}
    \item sample the next local state, observation and reward: $x_{t+1}, o_{t+1}, r_t \sim \simLocal(x_t, \ysrc_t, a_t)$
    \item sample the next RNN hidden state and influence source state : $z_{t+1}$, $\ysrc_{t+1} \sim \aip(\cdot|z_t, a_t, x_{t+1})$
\end{itemize}
The initial state $\SIALM_0$ of the IALS can be easily sampled by first sampling a full state $s \sim b_0$ and then extracting the local state and the influence source state $(x_0, \ysrc_0)$ from $s$.

\section{Experiments}
\label{sec:experiments}

We perform online planning experiments with the POMCP planner \citep{Silver2010} to answer the following questions:
when learning approximate influence predictors with RNNs, \vspace{-5pt}
\begin{itemize}
    \setlength{\itemsep}{0pt}
    \item can planning with an IALS be faster than planning with the global simulator while achieving similar performance, \textit{when the same number of simulations are allowed per planning step}?
    \item can planning with an IALS yield better performance than planning with the global simulator, \textit{when the same amount of planning time is allowed per planning step}?
\end{itemize}

\subsection*{Experimental Setup}
Our codebase was implemented in C++, including a POMCP planner and several benchmarking domains \footnote{available at \url{https://github.com/INFLUENCEorg/IAOP}}. We ran each of our experiments for many times on a computer cluster with the same amount of computational resources. To report results, we plot the means of evaluation metrics with standard errors as error bars. Details of our experiments are provided in the supplementary material.

\subsection*{Grab A Chair}
The first domain we use is the Grab A Chair domain mentioned in Section \ref{sec:Introduction}.  In our setting, the other agents employ a policy that selects chairs randomly in the beginning and greedily afterwards according to the frequency of observing to obtain a chair when visiting it.

Our intuition is that the amount of speedup we can achieve by replacing $\simGM$ with $\simIALMp$ depends on how fast we can sample influence source state variables $\Ysrc$ from the approximate influence predictor  $\aip$ and the size of hidden state variables $Y \backslash \Ysrc$ we can avoid simulating in $\simIALMp$. We perform planning with different simulators in games of $\{5,9,17,33,65,129\}$ agents for a horizon of $10$ steps, where a fixed number of $1000$ Monte Carlo simulations are performed per step.  

To obtain an approximate influence predictor $\aip$, we sample a dataset $\mathcal{D}$ of $1000$ episodes from the global simulator $\simGM$ with a uniform random policy and train a variant of RNN called Gated Recurrent Units (GRU) \citep{Cho2014} on $\mathcal{D}$ until convergence. To test if capturing the incoming influence is essential for achieving good performance when planning on $\simIALMp$, we use an IALS with a uniform random influence predictor as an additional baseline, denoted as $\simIALM^{\mathtt{random}}$.

\begin{figure}
    \centering
    \begin{subfigure}[b]{.49\textwidth}
        \centering
        \includegraphics[scale=0.29]{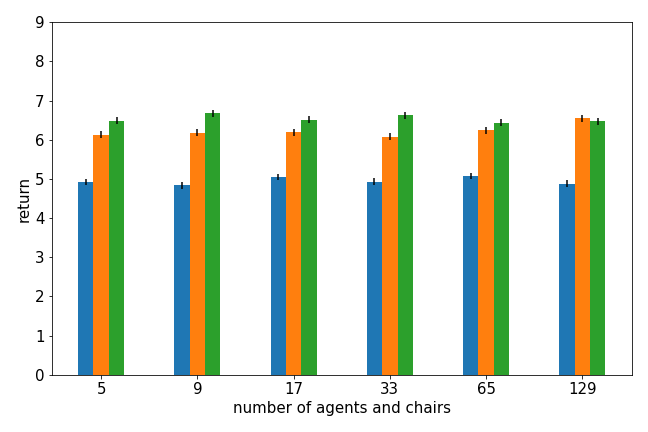}
        \caption{Average return}
        \label{fig:results-GAC-num_agents-return}
    \end{subfigure}
    \begin{subfigure}[b]{.49\textwidth}
        \centering
        \includegraphics[scale=0.29]{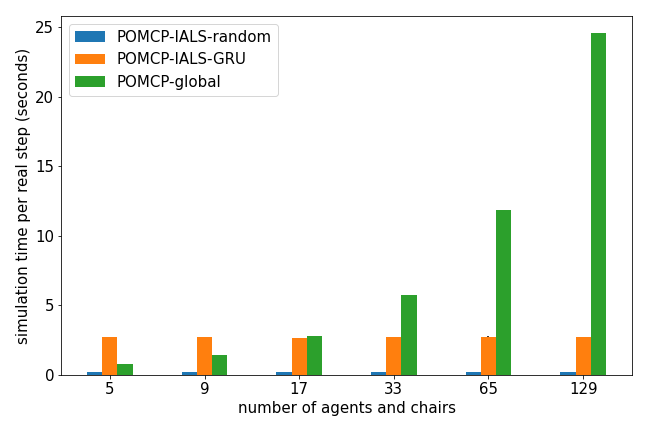}
        \caption{Average simulation time per step (seconds)}
        \label{fig:results-GAC-num_agents-time}
    \end{subfigure}
    \caption{Performance of POMCP with different simulators in Grab A Chair games of various sizes. While the IALS with GRU influence predictor achieves matching returns with the global simulator, the simulation is significantly faster in scenarios with many other agents.}
    \label{fig:results-GAC-num_agents}
\end{figure}

Figure \ref{fig:results-GAC-num_agents-return} shows the performance of planning with different simulators in scenarios of various sizes. It is clear that planning on $\simIALMp$ achieves significantly better performance than planning on $\simIALM^{\mathtt{random}}$, emphasizing the importance of learning  $\aip$ to capture the influence. While planning on $\simIALMp$ can indeed achieve matching performance with $\simGM$ as shown by the small differences in their returns, the advantage of the IALS, its speed, is shown in Figure \ref{fig:results-GAC-num_agents-time}. In contrast to $\simGM$ which slows down quickly because of the growing number of state variables to simulate, the computation time of both $\simIALMp$ and $\simIALM^{\mathtt{randomly}}$ barely increases. This is because those state variables added by more chairs and agents are abstracted away from the simulations in the IALS with their influence concisely captured by $\aip$ in the distribution of the two neighboring agents' decisions. Note that $\simIALMp$ is slower than $\simGM$ in scenarios with few agents due to the overheads of feedforward passing in the GRU. 

\begin{figure}
    \centering
    \includegraphics[scale=0.32]{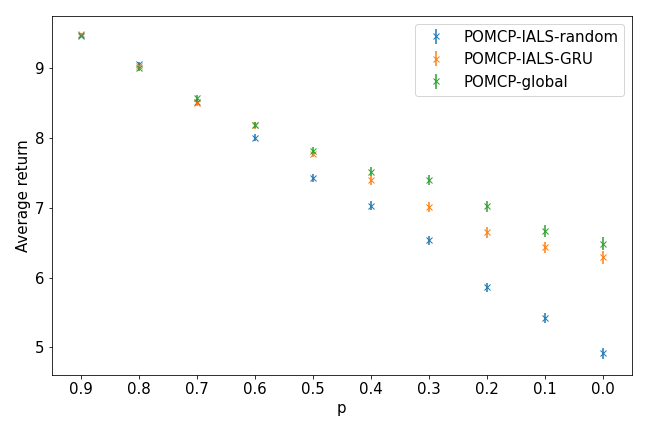}
    \caption{Performance of POMCP with different simulators in the modified Grab A Chair game under decreasing $p$, which implies stronger influence from the rest of the environment. The smaller performance difference between $\simIALMp$ and $\simGM$ under higher $p$ suggests that learning an accurate influence predictor is more important to achieve good planning performance when the local planning problem is more tightly coupled with the rest of the environment.}
    \label{fig:results-GAC-coupling}
\end{figure}

To further investigate how will influence-augmented online planning perform in environments with different \emph{influence strengths}, by which we mean the degree to which the local states are affected by the influence source states, we repeat our experiments above in a variant of the 5-agent Grab A Chair game where the only difference is that when two agents target the same chair, both of them have the same probability $p \in [0,1]$ to obtain the chair \footnote{Note that this leads to a physically unrealistic setting since it is possible that two agents obtain the same chair at a time step. However, it gives us a way to investigate the impact of the influence strength from the rest of the environment.}. The intuition is that when $p$ is lower, the influence from the rest of the environment will be stronger as the decisions of the two neighboring agents will be more decisive on whether the planning agent can secure a chair. In this case, higher prediction accuracy on the decisions of the two neighboring agents will be required for the agent to plan a good action. Figure \ref{fig:results-GAC-coupling} shows the planning performance with all simulators under decreasing $p$ which implies stronger influence strength from the rest of the environment. While the same amount of effort was put into training the approximate influence predictor $\aip$, the performance difference between planning with $\simIALMp$ and $\simGM$ is smaller under higher $p$. This suggests that in environments where the local planning problem is more tightly coupled with the rest of the environment, learning an accurate influence predictor $\aip$ is more important to achieve good planning performance.

\subsection*{Real-Time Online Planning in Grid Traffic Control}
The primary motivation of our approach is to improve online planning performance in realistic settings where the planning time per step is constrained. For this reason, we conduct real-time planning experiments in a more realistic domain called Grid Traffic Control, which simulates a busy traffic system with $9$ intersections, each of which consists of $4$ lanes with $6$ grids as shown in Figure \ref{fig:results-GTC-env}, with more details provided in the supplementary material.

The traffic lights are equipped with sensors providing $4$-bit information indicating if there are vehicles in the grids around them. While the other traffic lights employ a hand-coded switching strategy that prioritizes lanes with vehicles before the lights and without vehicles after the lights, the traffic light in the center is controlled by planning, with the goal to minimize the total number of vehicles in this intersection for a horizon of $30$ steps. 

As mentioned in Section \ref{background:POMCP}, POMCP approximates the belief update with an unweighted particle filter that reuses the simulations performed during the tree search. However, in our preliminary experiments, we observed the particle depletion problem, which occurred when POMCP ran out of particles because none of the existing particles was evidenced by the new observation. While to alleviate this problem we use a workaround inspired by \citet{Silver2010}  \footnote{While more advanced particle filters like Sequential Importance Resampling can reduce this problem, we chose to use POMCP in unmodified form to make it easier to interpret the benefits of our approach. Our workaround is that when the search tree is pruned because of a new observation, we add $N/6$ additional particles sampled from the initial belief $b_0$ to the current particle pool where $N$ is the number of remaining particles.}, when particle depletion still occurs at some point during an episode, the agent employs a uniform random policy.   

We train an influence predictor with a RNN and evaluate the performance of all three simulators $\simIALM^{\mathtt{random}}$, $\simIALMp$ and $\simGM$ in settings where the allowed planning time is fixed per step. Our hypothesis is that $\simIALMp$ will outperform the $\simGM$ when the planning time allowed is very constrained because in that case, the advantage on simulation speed will dominate the disadvantage on simulation accuracy caused by approximating the influence with $\aip$. 

Figure \ref{fig:results-GTC-real-time-num_sims} demonstrates the ability of the IALS to perform more than twice the number of simulations that can be performed by the global simulator within the same fixed time. This is directly translated into the ability of POMCP to plan for more time steps before the particle depletion occurs as shown in Figure \ref{fig:results-GTC-real-time-num_steps_to_go}. The more important effect of faster simulation is that our approach performs much better than planning on the global simulator especially when the planning time is limited. This suggests that there does exist a trade-off between simulation speed and simulation accuracy that allows planning on the IALS with an approximate influence predictor to achieve better online performance. 

Figure \ref{fig:results-GAC-real-time} in the supplementary material performs a similar time-constrained evaluation in the Grab A Chair domain. The finding there is that the advantage of the IALS on the simulation speed is clearer  when the global model of the problem is more complex, in which cases the IALS with an approximate influence predictor shows a superior performance compared to the global simulator.


\begin{figure}
    \centering
    \begin{subfigure}[b]{.49\textwidth}
        \centering
        \includegraphics[scale=0.14]{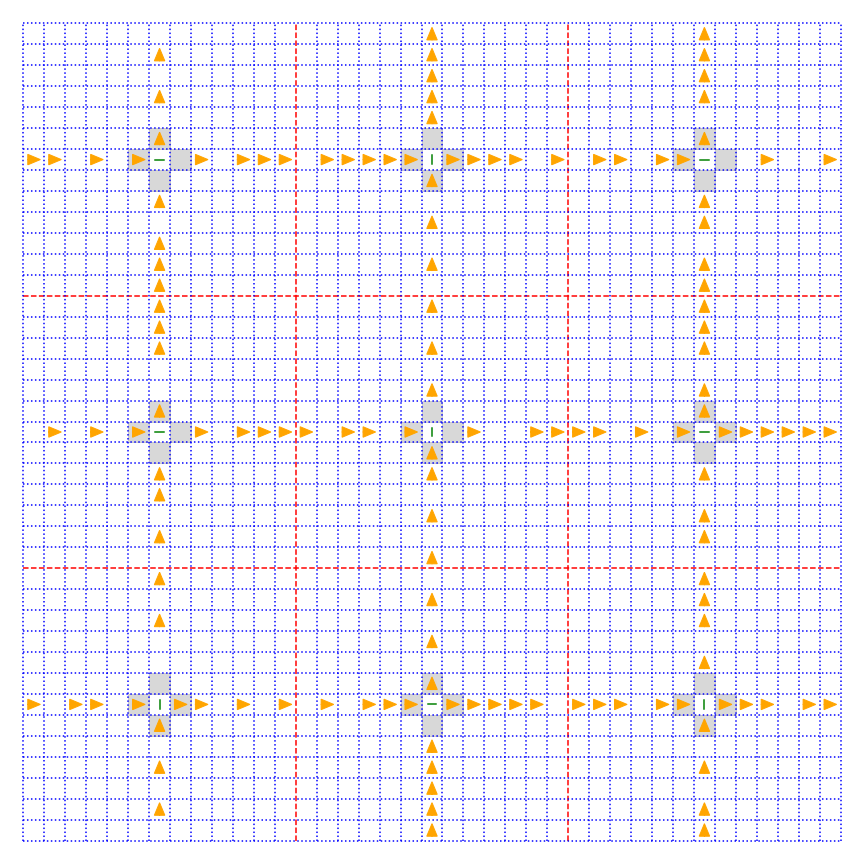}
        \caption{The Grid Traffic Control domain}
        \label{fig:results-GTC-env}
    \end{subfigure}
    \begin{subfigure}[b]{.49\textwidth}
        \centering
        \includegraphics[scale=0.27]{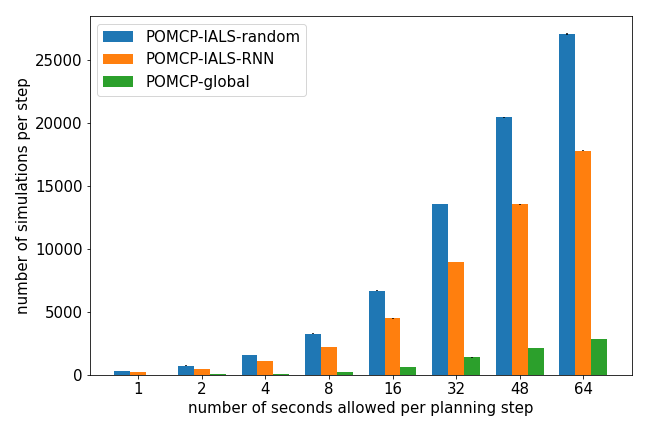}
        \caption{Number of simulations performed per planning step}
        \label{fig:results-GTC-real-time-num_sims}
    \end{subfigure}
    \begin{subfigure}[b]{.49\textwidth}
        \centering
        \includegraphics[scale=0.27]{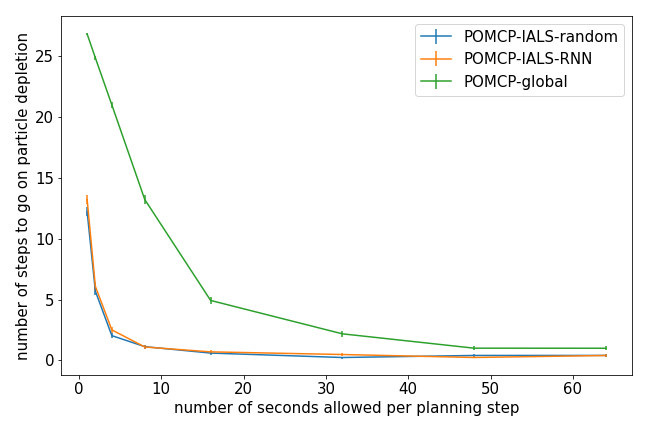}
        \caption{Number of steps to go on particle depletion}
        \label{fig:results-GTC-real-time-num_steps_to_go}
    \end{subfigure}
    \begin{subfigure}[b]{.49\textwidth}
        \centering
        \includegraphics[scale=0.27]{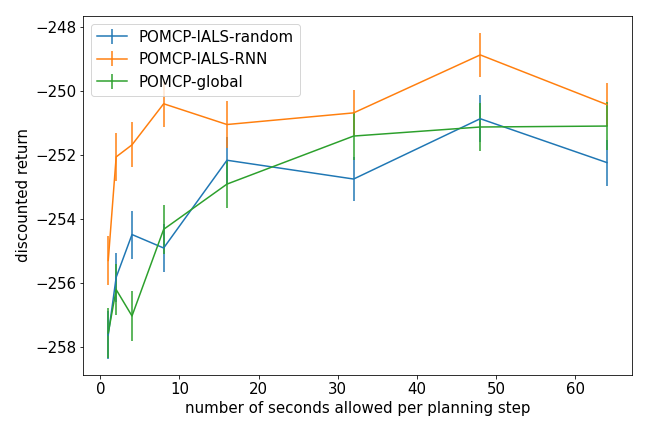}
        \caption{Discounted return}
        \label{fig:results-GTC-real-time-return}
    \end{subfigure}
    \caption{Performance of POMCP with different simulators while allowing different numbers of seconds per planning step in the Grid Traffic Control domain. While the planning performance of the IALS with trained influence predictor dominates the global simulator when the planning time is constrained, the performance difference decreases when more time is allowed.}
\end{figure}

\section{Related Work}
\label{sec:discussion}

The idea of 
utilizing offline knowledge learning for improved online planning performance 
has been well-studied
\citep{Gelly,Gelly2011,Silver2016,Silver2017,Silver2018,Anthony}.
These approaches can be categorized as 1)
learning value functions or policies to guide the tree search, 2) improving
default policy for more informative rollouts, 3) replacing rollouts with
learned value functions and 4) initializing state-action value estimates. Our
approach takes a distinct path by speeding up computationally expensive forward
simulations, which allows the planner to sample more trajectories for each
decision. 

Closest to our work is 
the approach by
\citet{chitnis2020learning}, which exploits
\emph{exogenous variables} to reduce the state space of the model for more efficient
simulation and planning. While both of the approaches learn a more compact model by
abstracting away state variables, exogenous variables are fundamentally
different from the non-local variables that we 
abstract away. 
By
definition, exogenous variables refer to those variables that are beyond the
control of the agent:
they cannot be affected, directly or indirectly, by the agent's actions
\citep{Boutilier1999,chitnis2020learning}. 
In contrast, the non-local variables that are abstracted away in IBA \citep{Oliehoek2012} 
can be chosen more freely, as long as they do not \emph{directly}
affect the agent's observation and reward. 
Therefore, the exogenous variables
and non-local variables are in general two different sets of variables that can
be exploited to reduce the state space size. 
For instance, in the traffic
problem of Figure \ref{fig:results-GTC-env}, there are no exogenous variables
as our action can directly or indirectly effect the transitions at other
intersections (by taking or sending vehicles from/to them). 
This demonstrates that our
approach allows us to reduce the state space of this problem beyond the
exogenous variables.

The idea of replacing a computationally demanding simulator with an approximate simulator for higher simulation efficiency has been explored in many fields under the name of \emph{surrogate model}, such as computer animation \citep{Grzeszczuk1999}, network simulation \citep{Kazer2018}, the simulation of seismic waves \citep{Moseley2018} and so on. Our work explores this idea in the context of sample-based planning in structured domains.

Recent works in deep model-based reinforcement learning \citep{Oh,Farquhar2018,Hafner2018,Schrittwieser2019,VanderPol2020} have proposed to learn an approximate model of the environment by interacting with it, and then plan a policy within the learned model for better sample efficiency. Our method considers a very different setting, in which we speed up the simulation for sample-based planning by approximating part of the global simulator, that is, the influence from the rest of the environment, and retain the simulation accuracy by explicitly utilizing a light and accurate local simulator.

\section{Conclusion}
\label{sec:conclusion}

In this work we aim to address the problem that simulators modeling the entire environment is often slow and hence not suitable for sample-based planning methods which require a vast number of Monte Carlo simulations to plan a good action. Our approach transforms an expensive factored global simulator into an influence-augmented local simulator (IALS) that is less accurate but much faster. The IALS utilizes a local simulator which accurately models the state variables that are most important to the planning agent and captures the influence from the rest of the environment with an approximate influence predictor learned offline. Our empirical results in the Grid Traffic Control domain show that in despite of the simulation inaccuracy caused by approximating the incoming influence with a recurrent neural network, planning on the IALS yields better online performance than planning on the global simulator due to the higher simulation efficiency, especially when the planning time per step is limited. While in this work we collect data from the global simulator with a random exploratory policy to learn the influence, a direction for future work is to study how this offline learning procedure can be improved for better performance during online planning. 

\clearpage
\section*{Broader Impact}

The potential impact of this work is precisely its motivation: making online planning more useful in real-world decision making scenarios, enabling more daily decisions to be made autonomously and intelligently, with promising applications including autonomous warehouse and traffic light control. 

Unlike simulators constructed by domain experts, which are in general easier to test and debug, influence-augmented local simulator contains an approximate influence predictor learned from data, which may fail with rare inputs and result in catastrophic consequences especially when controlling critical systems. This suggests that extensive testing and regulation will be required before deploying influence-augmented local simulators in real-world decision making scenarios. 

\begin{ack}
  \begin{wrapfigure}{r}{0.26\columnwidth}
    \centering
    \vspace{-20pt}
    \includegraphics[scale=0.1]{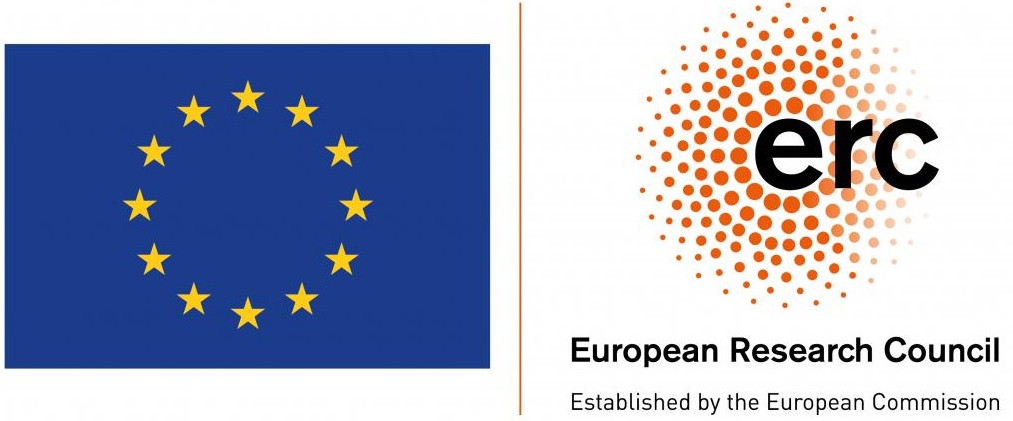}
\end{wrapfigure}
This project had received funding from the European Research Council (ERC) under the European Union's Horizon 2020 research and innovation programme (grant agreement No.~758824 \textemdash INFLUENCE).

\end{ack}

\small
\bibliographystyle{apalike}
\bibliography{references.bib}

\begin{thebibliography}{}

\bibitem[Anthony et~al., 2017]{Anthony}
Anthony, T., Tian, Z., and Barber, D. (2017).
\newblock {Thinking Fast and Slow with Deep Learning and Tree Search}.
\newblock In {\em Advances in Neural Information Processing Systems}, pages
  5360--5370.

\bibitem[Auer et~al., 2002]{Auer2002}
Auer, P., Cesa-Bianchi, N., and Fischer, P. (2002).
\newblock {Finite-time analysis of the multiarmed bandit problem}.
\newblock {\em Machine Learning}, 47(2-3):235--256.

\bibitem[Becker et~al., 2004]{Becker2004}
Becker, R., Zilberstein, S., and Lesser, V. (2004).
\newblock {Decentralized Markov decision processes with event-driven
  interactions}.
\newblock In {\em Proceedings of the Third International Joint Conference on
  Autonomous Agents and Multiagent Systems, AAMAS 2004}, volume~1, pages
  302--309.

\bibitem[Becker et~al., 2003]{Becker2003}
Becker, R., Zilberstein, S., Lesser, V., and Goldman, C.~V. (2003).
\newblock {Transition-Independent Decentralized Markov Decision Processes}.
\newblock In {\em Proceedings of the International Conference on Autonomous
  Agents}, volume~2, pages 41--48.

\bibitem[Boutilier et~al., 1999]{Boutilier1999}
Boutilier, C., Dean, T., and Hanks, S. (1999).
\newblock {Decision-Theoretic Planning: Structural Assumptions and
  Computational Leverage}.
\newblock {\em Journal of Artificial Intelligence Research}, 11:1--94.

\bibitem[Browne et~al., 2012]{Browne2012}
Browne, C.~B., Powley, E., Whitehouse, D., Lucas, S.~M., Cowling, P.~I.,
  Rohlfshagen, P., Tavener, S., Perez, D., Samothrakis, S., and Colton, S.
  (2012).
\newblock {A survey of Monte Carlo tree search methods}.

\bibitem[Chitnis and Lozano-P{\'e}rez, 2020]{chitnis2020learning}
Chitnis, R. and Lozano-P{\'e}rez, T. (2020).
\newblock Learning compact models for planning with exogenous processes.
\newblock In {\em Conference on Robot Learning}, pages 813--822. PMLR.

\bibitem[Cho et~al., 2014]{Cho2014}
Cho, K., {Van Merri{\"{e}}nboer}, B., Gulcehre, C., Bahdanau, D., Bougares, F.,
  Schwenk, H., and Bengio, Y. (2014).
\newblock {Learning phrase representations using RNN encoder-decoder for
  statistical machine translation}.
\newblock In {\em Conference on Empirical Methods in Natural Language
  Processing (EMNLP 2014)}, pages 1724--1734.

\bibitem[Coulom, 2006]{Coulom2006}
Coulom, R. (2006).
\newblock {Efficient selectivity and backup operators in Monte-Carlo tree
  search}.
\newblock In {\em International conference on computers and games}, volume 4630
  LNCS, pages 72--83.

\bibitem[Farquhar et~al., 2018]{Farquhar2018}
Farquhar, G., Rockt{\"{a}}schel, T., Igl, M., and Whiteson, S. (2018).
\newblock {TreeqN and ATreEC: Differentiable tree-structured models for deep
  reinforcement learning}.
\newblock In {\em 6th International Conference on Learning Representations,
  ICLR 2018 - Conference Track Proceedings}.

\bibitem[Gelly and Silver, 2007]{Gelly}
Gelly, S. and Silver, D. (2007).
\newblock {Combining online and offline knowledge in UCT}.
\newblock In {\em Proceedings of the 24th international conference on Machine
  learning}, volume 227, pages 273--280.

\bibitem[Gelly and Silver, 2011]{Gelly2011}
Gelly, S. and Silver, D. (2011).
\newblock {Monte-Carlo tree search and rapid action value estimation in
  computer Go}.
\newblock {\em Artificial Intelligence}, 175:1856--1875.

\bibitem[Grzeszczuk et~al., 1999]{Grzeszczuk1999}
Grzeszczuk, R., Terzopoulos, D., and Hinton, G. (1999).
\newblock {Fast neural network emulation of dynamical systems for computer
  animation}.
\newblock In {\em Advances in Neural Information Processing Systems}, pages
  882--888.

\bibitem[Hafner et~al., 2019]{Hafner2018}
Hafner, D., Lillicrap, T., Fischer, I., Villegas, R., Ha, D., Lee, H., and
  Davidson, J. (2019).
\newblock {Learning Latent Dynamics for Planning from Pixels}.
\newblock In {\em International Conference on Machine Learning}, pages
  2555----2565.

\bibitem[Hochreiter and Schmidhuber, 1997]{Hochreiter1997}
Hochreiter, S. and Schmidhuber, J. (1997).
\newblock {Long Short-Term Memory}.
\newblock {\em Neural Computation}, 9(8):1735--1780.

\bibitem[Kaelbling et~al., 1998]{Kaelbling1998}
Kaelbling, L.~P., Littman, M.~L., and Cassandra, A.~R. (1998).
\newblock {Planning and acting in partially observable stochastic domains}.
\newblock {\em Artificial Intelligence}, 101(1-2):99--134.

\bibitem[Kazer et~al., 2018]{Kazer2018}
Kazer, C.~W., Sedoc, J., Ng, K.~K., Liu, V., and Ungar, L.~H. (2018).
\newblock {Fast network simulation through approximation or: How blind men can
  describe elephants}.
\newblock In {\em HotNets 2018 - Proceedings of the 2018 ACM Workshop on Hot
  Topics in Networks}, pages 141--147.

\bibitem[Kearns et~al., 2002]{Kearns}
Kearns, M., Mansour, Y., and Ng, A.~Y. (2002).
\newblock {A sparse sampling algorithm for near-optimal planning in large
  Markov decision processes}.
\newblock {\em Machine learning}, 49(2-3):193----208.

\bibitem[Kocsis and Szepesv{\'{a}}ri, 2006]{Kocsis}
Kocsis, L. and Szepesv{\'{a}}ri, C. (2006).
\newblock {Bandit based Monte-Carlo planning}.
\newblock In {\em European conference on machine learning}, volume 4212 LNAI,
  pages 282--293.

\bibitem[Li et~al., 2006]{Li}
Li, L., Walsh, T.~J., and Littman, M.~L. (2006).
\newblock {Towards a unified theory of state abstraction for MDPs}.
\newblock In {\em 9th International Symposium on Artificial Intelligence and
  Mathematics, ISAIM 2006}.

\bibitem[Moseley et~al., 2018]{Moseley2018}
Moseley, B., Markham, A., and Nissen-Meyer, T. (2018).
\newblock {Fast approximate simulation of seismic waves with deep learning}.
\newblock {\em arXiv preprint arXiv:1807.06873}.

\bibitem[Oh et~al., 2017]{Oh}
Oh, J., Singh, S., and Lee, H. (2017).
\newblock {Value prediction network}.
\newblock In {\em Advances in Neural Information Processing Systems}, volume
  2017-Decem, pages 6119--6129.

\bibitem[Oliehoek et~al., 2019]{Oliehoek2019}
Oliehoek, F.~A., Witwicki, S., and Kaelbling, L.~P. (2019).
\newblock {A Sufficient Statistic for Influence in Structured Multiagent
  Environments}.
\newblock {\em arXiv preprint arXiv:1907.09278}.

\bibitem[Oliehoek et~al., 2012]{Oliehoek2012}
Oliehoek, F.~A., Witwicki, S.~J., and Kaelbling, L.~P. (2012).
\newblock {Influence-based abstraction for multiagent systems}.
\newblock In {\em Twenty-sixth AAAI conference on artificial intelligence}.

\bibitem[Petrik and Zilberstein, 2009]{Petrik2009}
Petrik, M. and Zilberstein, S. (2009).
\newblock {A Bilinear programming approach for multiagent planning}.
\newblock {\em Journal of Artificial Intelligence Research}, 35:235--274.

\bibitem[Ruder, 2016]{Ruder2016}
Ruder, S. (2016).
\newblock {An overview of gradient descent optimization algorithms}.
\newblock {\em arXiv preprint arXiv:1609.04747}.

\bibitem[Schrittwieser et~al., 2019]{Schrittwieser2019}
Schrittwieser, J., Antonoglou, I., Hubert, T., Simonyan, K., Sifre, L.,
  Schmitt, S., Guez, A., Lockhart, E., Hassabis, D., Graepel, T., Lillicrap,
  T., and Silver, D. (2019).
\newblock {Mastering Atari, Go, Chess and Shogi by Planning with a Learned
  Model}.
\newblock {\em arXiv preprint arXiv:1911.08265}.

\bibitem[Silver et~al., 2016]{Silver2016}
Silver, D., Huang, A., Maddison, C.~J., Guez, A., Sifre, L., {Van Den
  Driessche}, G., Schrittwieser, J., Antonoglou, I., Panneershelvam, V.,
  Lanctot, M., Dieleman, S., Grewe, D., Nham, J., Kalchbrenner, N., Sutskever,
  I., Lillicrap, T., Leach, M., Kavukcuoglu, K., Graepel, T., and Hassabis, D.
  (2016).
\newblock {Mastering the game of Go with deep neural networks and tree search}.
\newblock {\em Nature}, 529(7587):484--489.

\bibitem[Silver et~al., 2018]{Silver2018}
Silver, D., Hubert, T., Schrittwieser, J., Antonoglou, I., Lai, M., Guez, A.,
  Lanctot, M., Sifre, L., Kumaran, D., Graepel, T., Lillicrap, T., Simonyan,
  K., and Hassabis, D. (2018).
\newblock {A general reinforcement learning algorithm that masters chess,
  shogi, and Go through self-play}.
\newblock {\em Science}, 362(6419):1140--1144.

\bibitem[Silver et~al., 2017]{Silver2017}
Silver, D., Schrittwieser, J., Simonyan, K., Antonoglou, I., Huang, A., Guez,
  A., Hubert, T., Baker, L., Lai, M., Bolton, A., Chen, Y., Lillicrap, T., Hui,
  F., Sifre, L., {Van Den Driessche}, G., Graepel, T., and Hassabis, D. (2017).
\newblock {Mastering the game of Go without human knowledge}.
\newblock {\em Nature}, 550(7676):354--359.

\bibitem[Silver and Veness, 2010]{Silver2010}
Silver, D. and Veness, J. (2010).
\newblock {Monte-Carlo planning in large POMDPs}.
\newblock In {\em Advances in neural information processing systems}, pages
  2164----2172.

\bibitem[Suau et~al., 2020]{suau2019influence}
Suau, M., Congeduti, E., Starre, R., Czechowski, A., and Olihoek, F. (2020).
\newblock Influence-aware memory for deep reinforcement learning.
\newblock {\em arXiv preprint arXiv:1911.07643}.

\bibitem[Van~der Pol et~al., 2020]{VanderPol2020}
Van~der Pol, E., Kipf, T., Oliehoek, F.~A., and Welling, M. (2020).
\newblock {Plannable Approximations to MDP Homomorphisms: Equivariance under
  Actions}.
\newblock In {\em Proceedings of the 19th International Conference on
  Autonomous Agents and MultiAgent Systems}, pages 1431----1439.

\bibitem[Witwicki and Durfee, 2010]{Witwicki2010}
Witwicki, S.~J. and Durfee, E.~H. (2010).
\newblock {Influence-based policy abstraction for weakly-coupled Dec-POMDPs}.
\newblock In {\em ICAPS 2010 - Proceedings of the 20th International Conference
  on Automated Planning and Scheduling}, pages 185--192.

\end{thebibliography}

\clearpage
\section{Supplementary Material}
\label{sec:appnedix}

In this supplementary material, we provide the details of our experimental setups for reproducibility.

\subsection{Grab A Chair}

\subsubsection*{Environment}

The Grab A Chair game is an $N$-agent game where at every time step, each agent has an action space of two, trying to grab the chair on its left or right side. An agent only secures a chair if its targeted chair is not targeted by a neighboring agent. At the end of a time step $t$, each agent with $s_{t+1}{\in}\{0,1\}$ indicating if this agent obtains a chair receives a reward $r_t{=}s_{t+1}$ and a noisy observation $o_{t+1}$ on $s_{t+1}$ which has a probability $0.2$ to be flipped. 

In the experiments of Figure \ref{fig:results-GAC-coupling}, when two agents target the same chair, both of them have a probability of $p \in [0,1]$ to secure the chair, which means that there is a probability that two neighboring agents obtain the same chair. The following setup applies to all the experiments in this domain.

\subsubsection*{Experimental Setup}

\textbf{Influence Learning}

\begin{figure}[H]
    \centering
    \begin{subfigure}[b]{.32\textwidth}
        \centering
        \includegraphics[scale=0.22]{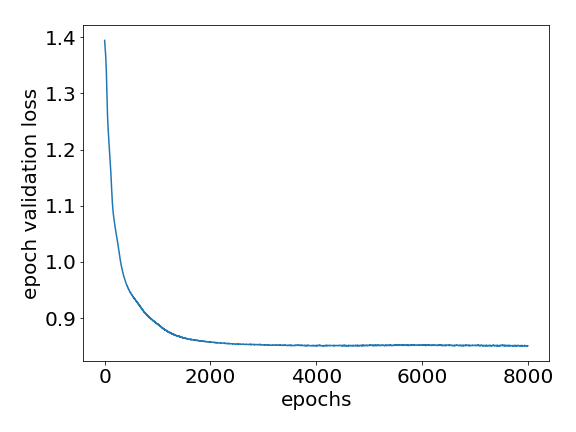}
        \caption{5 agents}
    \end{subfigure}
    \begin{subfigure}[b]{.32\textwidth}
        \centering
        \includegraphics[scale=0.22]{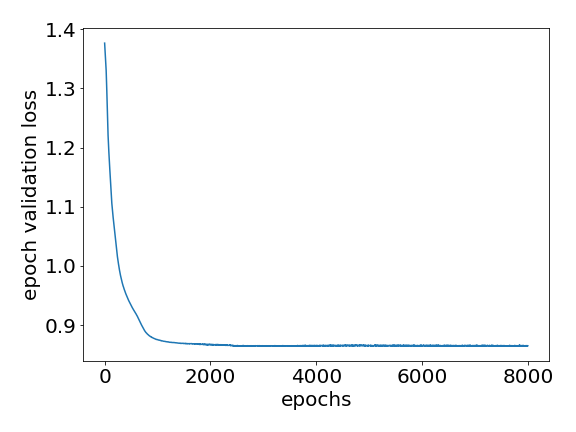}
        \caption{9 agents}
    \end{subfigure}
    \begin{subfigure}[b]{.32\textwidth}
        \centering
        \includegraphics[scale=0.22]{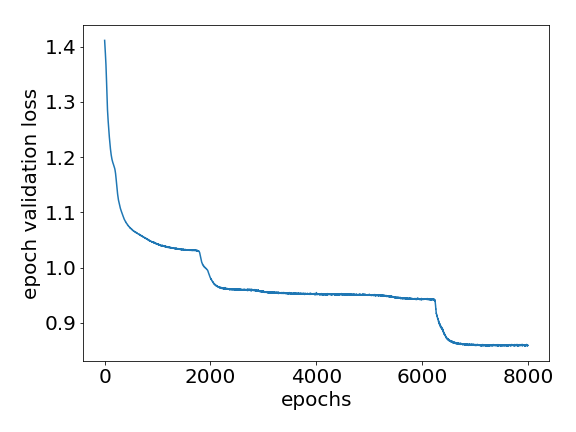}
        \caption{17 agents}
    \end{subfigure}
    \begin{subfigure}[b]{.32\textwidth}
        \centering
        \includegraphics[scale=0.22]{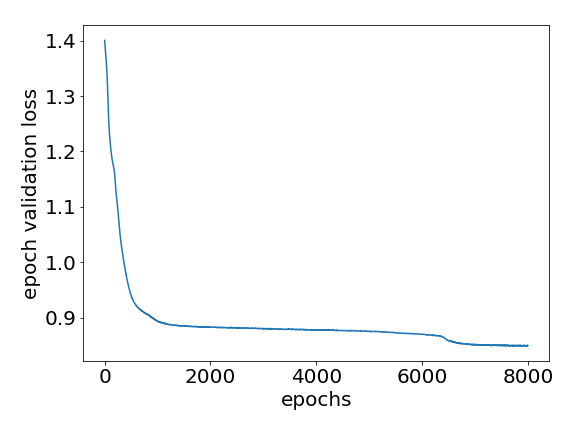}
        \caption{33 agents}
    \end{subfigure}
    \begin{subfigure}[b]{.32\textwidth}
        \centering
        \includegraphics[scale=0.22]{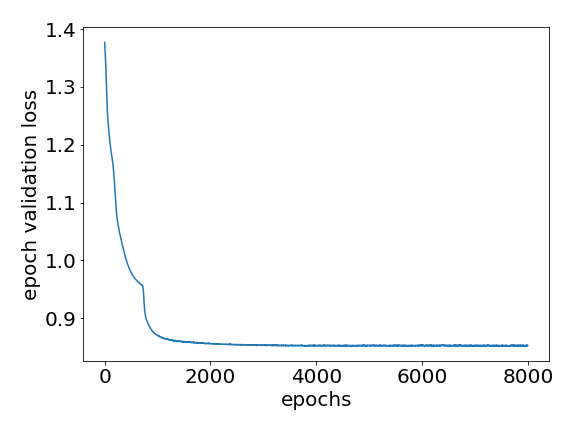}
        \caption{65 agents}
    \end{subfigure}
    \begin{subfigure}[b]{.32\textwidth}
        \centering
        \includegraphics[scale=0.22]{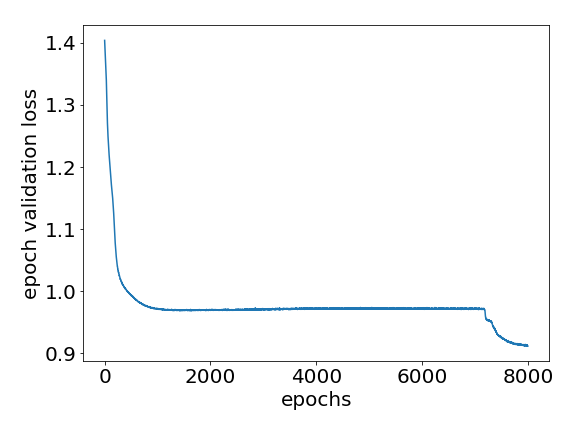}
        \caption{129 agents}
    \end{subfigure}
    \caption{Learning curves of influence predictors in the Grab A Chair domain.}
    \label{learning_curves:gac}
\end{figure}

In this domain, the approximate influence predictor $\aip$ is parameterized by a GRU \citep{Cho2014} classifier with $8$ hidden units. The dataset $\D$ consists of $1000$ episodes collected from the global simulator $\simGM$ with a uniform random policy, where $800$ episodes are used as the training set and the other $200$ episodes are used as the validation set. The hyperparameters used to train the GRU influence predictors in scenarios with $\{5, 9, 17, 33, 65, 129\}$ agents are shown in Table \ref{hyperparameters:GAC} and their learning curves are shown in Figure \ref{learning_curves:gac}.

\begin{table}[H]
    \caption{Hyperparameters used to train the GRU influence predictors for experiments in the Grab A Chair domain, where the weight decay was fine tuned within the range until there is no clear sign of overfitting.}
    \label{hyperparameters:GAC}
    \centering
    \begin{tabular}{@{}llr@{}} \toprule
        \centering
        Learning rate & $0.0005$  \\ 
        Batch size & 128  \\ 
        Number of epochs & 8000 \\
        Weight decay & $[1 \times 10^{-5}, 5 \times 10^{-5}]$  \\  \bottomrule
    \end{tabular}
  \end{table}

\clearpage
\textbf{Planning with POMCP}

The parameters used in the planning experiments with POMCP are shown in Table \ref{hyperparameters:GAC-planning}. 
\begin{table}[H]
    \centering
    \caption{Parameters for the planning experiments with POMCP in the Grab A Chair domain.}
    \label{hyperparameters:GAC-planning}
    \begin{tabular}{@{}llr@{}} \toprule
     Discount factor & $1.0$  \\ 
     Horizon & $10$  \\
     Number of simulations per step & $1000$  \\ 
     Number of initial particles & $1000$ \\
     Exploration constant in the UCB1 algorithm ($c$) & $100.0$  \\ \bottomrule
    \end{tabular}
\end{table}

\textbf{Real-time Online Planning in Grab A Chair domain}

\begin{figure}
    \centering
    \begin{subfigure}[b]{.325\textwidth}
        \centering
        \includegraphics[scale=0.195]{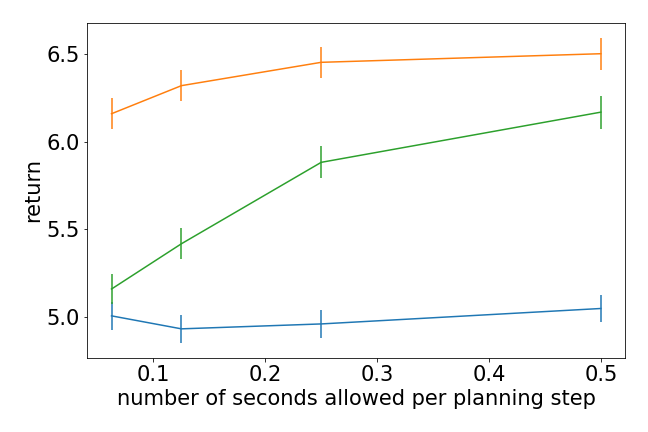}
    \end{subfigure}
    \begin{subfigure}[b]{.325\textwidth}
        \centering
        \includegraphics[scale=0.195]{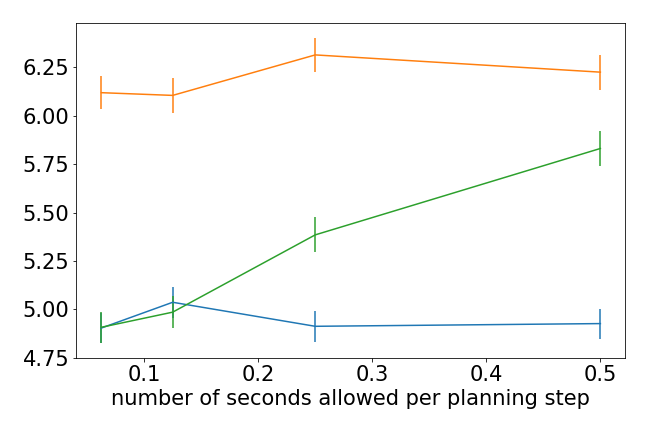}
    \end{subfigure}
    \begin{subfigure}[b]{.325\textwidth}
        \centering
        \includegraphics[scale=0.195]{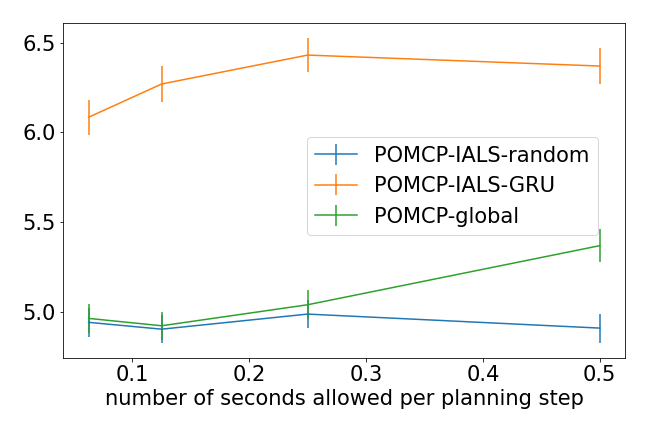}
    \end{subfigure}
    \begin{subfigure}[b]{.325\textwidth}
        \centering
        \includegraphics[scale=0.195]{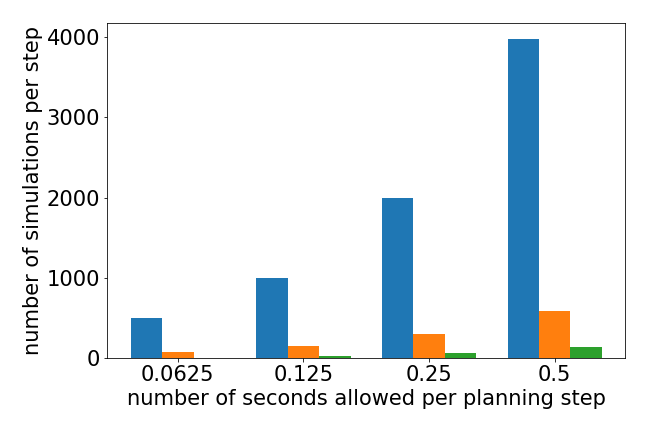}
    \end{subfigure}
    \begin{subfigure}[b]{.325\textwidth}
        \centering
        \includegraphics[scale=0.195]{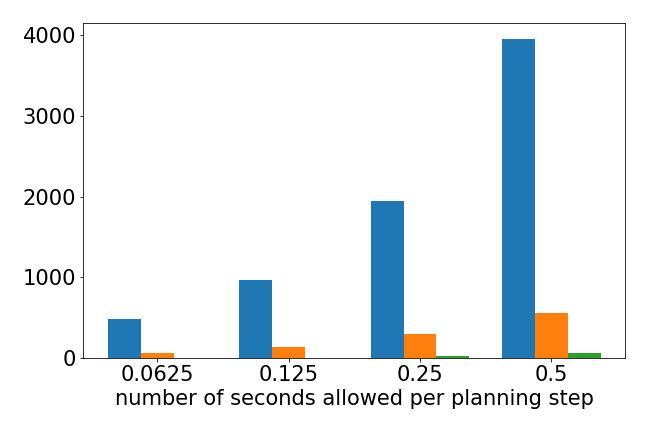}
    \end{subfigure}
    \begin{subfigure}[b]{.325\textwidth}
        \centering
        \includegraphics[scale=0.195]{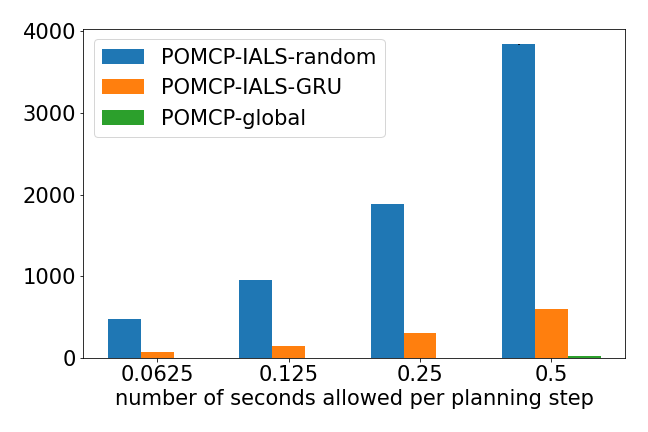}
    \end{subfigure}
    \caption{Performance of POMCP with different simulators while allowing different numbers of seconds per planning step in Grab A Chair games with 33 (left), 65 (middle) and 129 (right) agents. The advantage of IALS with GRU influence predictor over global simulator becomes clearer as the global model of the planning problem gets more complex.}
    \label{fig:results-GAC-real-time}
\end{figure}

We conduct a time-constrained evaluation in this domain with $\{33, 65,129\}$ agents, similar to the one performed in the Grid Traffic Control domain, where different amount of time is allowed to plan an action. Results in Figure \ref{fig:results-GAC-real-time} show that the advantage of the IALS with GRU influence predictor is clearer when the global model of the planning gets more complex.

\subsection{Grid Traffic Control}

\subsubsection*{Environment}

The Grid Traffic Control environment simulates a traffic system of $9$ intersections as shown in Figure \ref{fig:results-GTC-env}. The vehicles, plotted as yellow arrows, move from the left to right and the bottom to top, governed by the traffic lights in the center of each intersection. While they are initially generated with a probability of $0.7$ in each grid, new vehicles will enter the traffic system at entrances on the left and bottom borders whenever they are not occupied at last time step. When reaching the right and bottom borders, with a probability of $0.3$, vehicles leave the traffic system. 

While the other traffic lights are controlled by fixed switching strategies, the traffic light in the center intersection is controlled by the planning agent, whose action space consists of actions to set the light green for each lane. After an action $a_t$ is taken which results in the movement of vehicles, the agent receives an observation consisting of four Boolean variables $o_{t+1}{=}\{ \mathtt{left\_occupied}, \mathtt{right\_occupied}, \mathtt{up\_occupied}, \mathtt{bottom\_occupied}\}$ indicating if the four grids around the traffic light are occupied. The reward $r_t$ is the negative number of grids that are occupied in this intersection after the transition at time step $t$.

\subsubsection*{Experimental Setup}

\textbf{Influence Learning}

\begin{figure}
    \centering
    \includegraphics[scale=0.3]{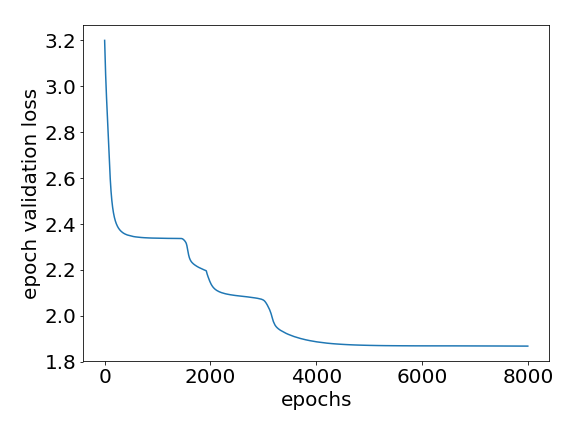}
    \caption{The learning curve of the influence predictor in the Grid Traffic Control domain.}
    \label{learning_curves:gtc}
\end{figure}

In this domain, the approximate influence predictor $\aip$ is parameterized by a RNN classifier with $2$ hidden units. The dataset $\D$ consists of $1000$ episodes collected from the global simulator $\simGM$ with a uniform random policy, where $800$ episodes are used as the training set and the other $200$ episodes are used as the validation set. The hyperparameters used to train the RNN influence predictor are shown in Table \ref{hyperparameters:GTC} and its learning curve is shown in Figure \ref{learning_curves:gtc}.

\begin{table}[H]
    \centering
    \caption{Hyperparameters used to train the RNN influence predictor for experiments in the Grid Traffic Control domain.}
    \vspace{3pt}
    \label{hyperparameters:GTC}
    \begin{tabular}{ |c|c| } 
     \hline
     Learning rate & $0.00025$  \\ 
     Batch size & $128$  \\ 
     Number of epochs & $8000$ \\
     Weight decay & $1 \times 10^{-4}$  \\ 
     \hline
    \end{tabular}
\end{table}

\textbf{Planning with POMCP}

The parameters used in the planning experiments with POMCP are shown in Table \ref{hyperparameters:GTC-planning}, where effective horizon is the maximal depth from the root node that a search or a rollout will be performed.
\begin{table}[H]
    \centering
    \caption{Parameters for the planning experiments with POMCP in the Grid Traffic Control domain.}
    \vspace{3pt}
    \label{hyperparameters:GTC-planning}
    \begin{tabular}{ |c|c| } 
     \hline
     Discount factor & $0.95$  \\ 
     Horizon & $30$  \\
     Number of seconds allowed per planning step & $\{1,2,4,8,16,32,48,64\}$  \\ 
     Number of initial particles & $1000$ \\
     Exploration constant in the UCB1 algorithm ($c$) & $10.0$  \\ 
     Effective horizon & $18$ \\
     \hline
    \end{tabular}
\end{table}

\end{document}